\pdfoutput=1

\documentclass[11pt]{article}

\usepackage[preprint]{acl}

\usepackage{times}
\usepackage{latexsym}
\usepackage{comment}
\usepackage[T1]{fontenc}

\usepackage[utf8]{inputenc}

\usepackage{microtype}

\usepackage{inconsolata}

\usepackage{graphicx}

\usepackage{algorithm}
\usepackage{algpseudocode}
\usepackage{amsmath}
\usepackage{booktabs}
\usepackage{multirow}
\newcommand{\ourmethod}{\textsc{AutoCT}}

\usepackage{tcolorbox}
\tcbuselibrary{listingsutf8}
\usepackage{listings}
\usepackage{caption}
\usepackage{multicol}
\usepackage{enumitem}

\title{\ourmethod: Automating Interpretable Clinical Trial Prediction with LLM Agents}

\author{
 \textbf{Fengze Liu\textsuperscript{1}},
 \textbf{Haoyu Wang\textsuperscript{1}},
 \textbf{Joonhyuk Cho\textsuperscript{2}},
 \textbf{Dan Roth\textsuperscript{1}},
 \textbf{Andrew W. Lo\textsuperscript{2}}
\\
 \textsuperscript{1}University of Pennsylvania \\
 \textsuperscript{2}Massachusetts Institute of Technology 
}

\begin{document}
\maketitle
\begin{abstract}
Clinical trials are critical for advancing medical treatments but remain prohibitively expensive and time-consuming. Accurate prediction of clinical trial outcomes can significantly reduce research and development costs and accelerate drug discovery. While recent deep learning models have shown promise by leveraging unstructured data, their black-box nature, lack of interpretability, and vulnerability to label leakage limit their practical use in high-stakes biomedical contexts. In this work, we propose \ourmethod, a novel framework that combines the reasoning capabilities of large language models with the explainability of classical machine learning. \ourmethod\ autonomously generates, evaluates, and refines tabular features based on public information without human input. Our method uses Monte Carlo Tree Search to iteratively optimize predictive performance. Experimental results show that \ourmethod\ performs on par with or better than SOTA methods on clinical trial prediction tasks within only a limited number of self-refinement iterations, establishing a new paradigm for scalable, interpretable, and cost-efficient clinical trial prediction. 
\end{abstract}

\section{Introduction}
Clinical trials are essential to bring new treatments to patients, yet they are extremely costly in terms of both time and resources. The process of developing a new drug and bringing it to market takes, on average, over 10 years and costs nearly US\$2.6 billion, with a clinical success rate of less than 10\% \cite{qian2025enhancing}. This long development cycle, combined with high costs and low success rates, poses significant challenges to drug discovery and development \cite{lo2022healthcare}. Accurate and informative predictions of clinical trial outcomes—and other related indicators—have the potential to guide clinical research, reduce costs, and accelerate the drug development pipeline.

Early studies apply classical machine learning models to predict trial outcomes using expert-curated features \cite{gayvert2016data, lo2019machine, siah2021predicting}. Although these approaches achieve robust performance, they are limited by their reliance on manually annotated tabular data. Such methods are not well-suited to incorporate unstructured information from diverse biomedical databases. 

More recent work has employed deep learning models capable of leveraging idiosyncratic data from various sources \cite{fu2022hint, wang2023spot, yue2024clinicalagent}. These models integrate information such as disease hierarchies, similarities with prior trials, drug toxicity profiles, and trial design attributes. While powerful, these deep learning frameworks often function as “black boxes,” making their predictions difficult to interpret. In high-stakes domains such as clinical trials, where interpretability and uncertainty quantification are critical, this lack of transparency can limit their practical applicability. Furthermore, many existing deep learning or embedding-based approaches risk label leakage when extracting information from external databases without enforcing a knowledge cutoff \cite{fu2022hint}.

\begin{figure*}[t]
    \centering
    \vspace{-0.5em}
    \includegraphics[width=\linewidth]{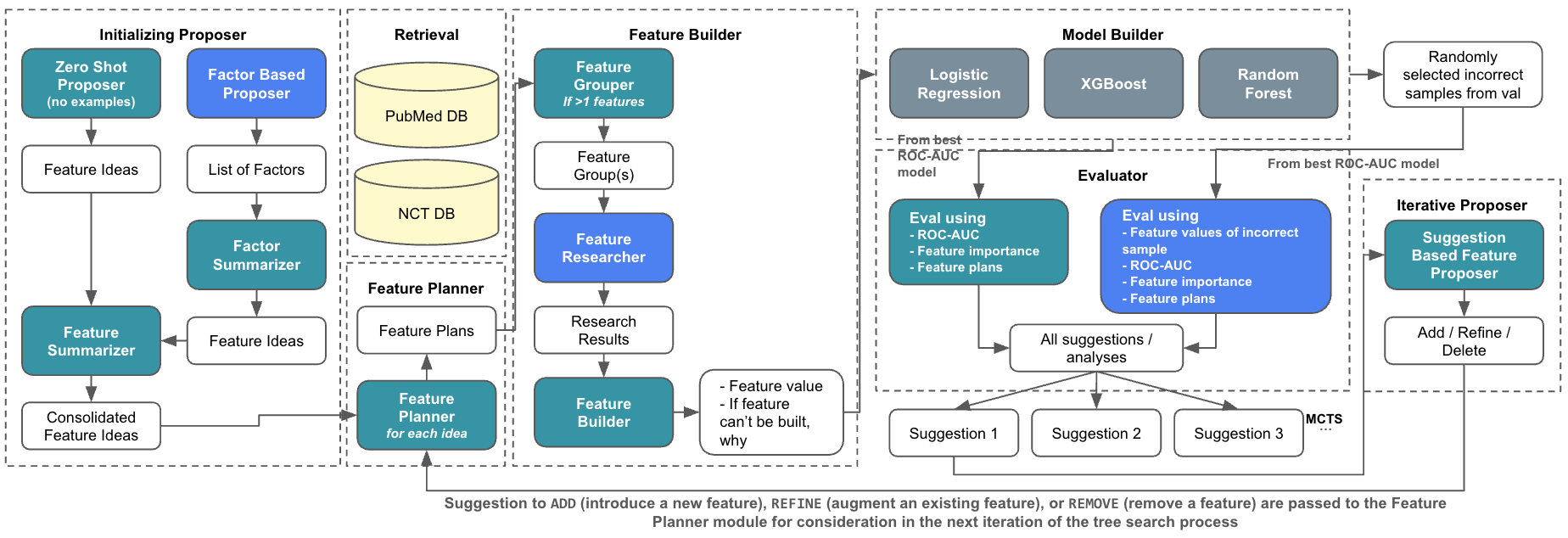}
    \caption{Overview of the \ourmethod\ Framework. Turquoise boxes indicate components using LLMs with Chain-of-Thought (CoT) reasoning. Blue boxes represent components using LLMs with ReAct-style reasoning (interleaving reasoning and action). White boxes denote inputs and outputs, while gray boxes correspond to standard function calls without LLM involvement.}
    \vspace{-1em}
    \label{fig:framework}
\end{figure*}

In this work, we introduce \textbf{\ourmethod}\ (\textbf{Auto}mated Interpretable \textbf{C}linical \textbf{T}rial Prediction with LLM Agents), a framework that addresses these limitations by combining the reasoning abilities of large language models (LLMs) with the interpretability and reliability of classical machine learning. \ourmethod\ is motivated by a few key observations. First, LLMs encode extensive prior knowledge in their parameters \cite{zhou2024conceptual, li-etal-2024-deceptive}, which can be harnessed in place of domain experts to suggest potentially predictive features and guide the feature engineering process. Second, LLMs has demonstrated certain reasoning abilities \cite{wei2022chain, yao2023react}, which enhance their test-time compute and allow them to perform complex tasks such as researching, planning, and constructing higher-level features. Third, existing generalized AutoML frameworks often rely on fixed pipelines or, when augmented with LLMs, provide feedback based only on limited information such as performance metrics and prior knowledge. In contrast, expert-driven approaches to clinical trial prediction typically involve in-depth analysis, contextual interpretation, and iterative refinement grounded in domain-specific research.

To better emulate this expert workflow, \ourmethod\ leverages LLMs not simply for guidance but as active agents in the feature construction process. As shown in Figure~\ref{fig:framework}, the system autonomously proposes, plans, and builds tabular features with extensive research. These features are then used to train a classical machine learning model. The resulting model performance is evaluated by an LLM, which conducts error analysis and offers iterative suggestions for improvement. Model performance is optimized via Monte Carlo Tree Search \cite{coulom2006efficient}, guided by these LLM-generated recommendations. Given only a clinical trial identifier and an outcome label (e.g., binary success or failure), \ourmethod\ can achieve performance competitive with state-of-the-art methods —without human intervention and within a limited number of MCTS iterations.

The contributions of this paper are threefold:
\begin{itemize}
\setlength\itemsep{-0.3em}
\item We present a novel and generalizable framework that achieves competitive performance on clinical classification tasks with SOTA methods. 
\item To the best of our knowledge, \ourmethod\ is the first end-to-end clinical machine learning framework that automatically extracts features with the help of LLMs, and is capable of dynamic feature discovery based on the scope of accessible data sources without human intervention. 
\item Our approach combines the interpretability of classical machine learning with the reasoning capabilities of LLMs, enabling transparent and quantifiable predictions suitable for high-stakes clinical decision-making. 
\vspace{-0.3em}
\end{itemize}


\section{Related Work}

\subsection{Clinical Trial Outcome Prediction}

Our work targets interpretable and quantifiable prediction of clinical trial outcomes. Early studies used classical machine learning models on expert-curated tabular features. For example, \citet{lo2019machine} showed strong performance on Phase 2-to-approval (P2APP) and Phase 3-to-approval (P3APP) datasets using imputation and models like random forests and SVMs, while \citet{siah2021predicting} improved results through a domain-informed data science challenge with Novartis. However, these methods struggle to incorporate unstructured or multi-modal data due to their reliance on fixed tabular inputs.

More recent work employs deep learning to address these limitations. \citet{fu2022hint} proposed HINT, a GNN model that integrates multiple data sources, and \citet{wang2023spot} introduced a meta-learning framework that leverages temporal trial sequences. While effective, such models act as black boxes with limited interpretability. \citet{yue2024clinicalagent} presented ClinicalAgent, a multi-agent system that enhances transparency via external tools, though LLMs still face issues with calibrated uncertainty and label leakage \cite{xiong2023can, li-etal-2024-deceptive, fu2022hint}.

\ourmethod\ bridges these paradigms by using LLMs solely for feature construction, with interpretable classical models for prediction. This enables multi-source integration while retaining transparency and robustness.

\subsection{Automated Machine Learning (AutoML)}

Most existing AutoML approaches begin with a predefined tabular dataset comprising engineered features and corresponding labels. \citet{han2024large} employed LLMs to generate additional feature rules based on existing tabular inputs, thereby enhancing the performance of downstream predictive models. \citet{hollmann2023caafe, li2025exploring} leveraged LLMs' prior knowledge for feature engineering. \citet{luo2024autom3l} proposed embedding LLMs at each stage of the machine learning workflow. \citet{chi2024sela} introduced an AutoML framework that uses Monte Carlo Tree Search to optimize the entire machine learning pipeline.

Inspired by these works, our approach aims to further extend automation by removing the dependency on an initial feature set. Instead, \ourmethod\ uses the contextual understanding and reasoning capabilities of LLMs to propose and construct an initial set of tabular features from only the system prompt, the unique identifier of a trial and the target label. These features are then iteratively refined and optimized using MCTS, enabling fully automated, end-to-end clinical trial modeling.

\begin{figure*}[t]
\small
\centering
\begin{minipage}{\textwidth}
\vspace{-0.5em}
\begin{algorithm}[H]
\caption{\ourmethod\ with MCTS}\label{mcts}
\begin{algorithmic}[1]
\Require Train/validation sets $\mathcal{I}_t$, $\mathcal{I}_v$, labels $\mathcal{Y}_t$, $\mathcal{Y}_v$, task description $\mathcal{D}$, iterations $N$
\State $\mathcal{F}_0 \gets$ \textbf{InitializingProposer}$(\mathcal{D}, \mathcal{I}_t', \mathcal{Y}_t')$, where $\mathcal{I}_t' \subset \mathcal{I}_t$, $\mathcal{Y}_t' \subset \mathcal{Y}_t$
\State $F_0 \gets$ \textbf{FeatureBuilder}($\mathcal{I}_t$, \textbf{FeaturePlanner}$(\mathcal{F}_0))$
\State $M_0 \gets$ \textbf{ModelBuilder}$(F_0, \mathcal{Y}_t)$, score $s_0 \gets$ \textbf{Evaluate}$(M_0, \mathcal{I}_v, \mathcal{Y}_v)$
\State Initialize tree $\mathcal{T}$ with root $(\mathcal{F}_0, s_0)$
\For{$i = 1$ to $N$}
    \For{$g \in$ \textbf{Evaluator}$(s_{i-1}, F_{i-1}, \mathcal{I}_v', \mathcal{Y}_v', \mathcal{D})$}, where $\mathcal{I}_v' \subset \mathcal{I}_v$, $\mathcal{Y}_v' \subset \mathcal{Y}_v$
        \State $F' \gets$ \textbf{UpdateFeatures}$(F_{i-1},$ \textbf{FeatureBuilder}($\mathcal{I}_t$, \textbf{FeaturePlanner}(\textbf{IterativeProposer}$(g))))$
        \State $M' \gets$ \textbf{ModelBuilder}$(F', \mathcal{Y}_t)$, score $s' \gets$ \textbf{Evaluate}$(M', \mathcal{I}_v, \mathcal{Y}_v)$
        \State Add $(F', s')$ to $\mathcal{T}$
    \EndFor
    \State $(F_i, s_i) \gets$ \textbf{SelectBestChild}$(\mathcal{T})$
\EndFor
\State \Return Best feature set/model from $\mathcal{T}$
\end{algorithmic}
\end{algorithm}
\vspace{-1em}
\end{minipage}
\caption{\ourmethod\ algorithm using Monte Carlo Tree Search to iteratively refine features and improve prediction.}
\label{fig:alg}
\end{figure*}

\subsection{LLM Agents in Healthcare}
Large Language Model (LLM)-based agents have emerged as transformative tools in the healthcare domain \cite{wang2025survey}. Systems like MedAide \cite{wei2024medaide} coordinate agents across stages of diagnosis and treatment, while frameworks such as MDAgents~\cite{kim2024mdagents} adaptively assign collaboration structures to teams of LLMs. Agent Hospital~\cite{li2024agent} simulates entire hospital environments with LLM-powered agents, and \citet{dutta2024adaptive} proposed a simulated doctor-patient dialogue agent to enhance diagnostic reasoning, which showed strong performance on benchmarks such as MedQA. 

While existing LLM agents focus on tasks like diagnosis and documentation, \ourmethod\ introduces LLM agents for feature discovery in clinical trial prediction. By simulating a full machine learning pipeline using planning, example-based reasoning and model feedback, \ourmethod\ bridges the gap between LLM-based reasoning and structured machine learning in the biomedical domain.



\section{Methods}

We begin by introducing the retrieval tools available to our agents for steps that require in-depth reasoning and research (\S~\ref{sec:retrieval_tools}). The \ourmethod\ framework consists of several key components: the Feature Proposer (\S~\ref{sec:feature_proposer}), which generates conceptual feature ideas grounded in both parametric knowledge and selected training samples; the Feature Planner (\S~\ref{sec:feature_planner}), which transforms these ideas into executable instructions and structured schemas; the Feature Builder (\S~\ref{sec:feature_builder}), which conducts external research and computes the corresponding feature values; the Model Builder (\S~\ref{sec:model_builder}), which trains classical machine learning models on the derived features; and the Evaluator (\S~\ref{sec:evaluator}), which assesses model performance and provides iterative feedback. These components interact within a Monte Carlo Tree Search framework (\S~\ref{sec:mcts}), where each Evaluator suggestion constitutes a new node to refine the feature space. The full algorithm is illustrated in Figure~\ref{fig:alg}. To improve reasoning efficiency and performance in complex modules—specifically the Feature Proposer, Feature Builder, and Evaluator—we adopt a multi-agent architecture. This design decomposes each complex task into smaller, more manageable sub-tasks, enabling us to provide each LLM with shorter and more targeted prompts. This hierarchical reasoning framework also explicitly encourages deeper thinking from each LLM to achieving robust performance. 

\subsection{Retrieval Tools}
\label{sec:retrieval_tools}
To replicate the research process of biomedical experts in clinical trial outcome prediction, we equip LLMs with tools to conduct external research during steps that require more complex reasoning, similar in spirit to retrieval-augmented generation (RAG) \cite{lewis2020retrieval}. To support this, we embed academic articles from PubMed \cite{white2020pubmed} and clinical trial records from ClinicalTrials.gov \cite{zarin2011clinicaltrials} using PubMedBERT-based embeddings \cite{mezzetti2023embeddings}, creating two local knowledge bases referred to as PubMed DB and NCT DB, respectively, as shown in Figure~\ref{fig:framework}.

When the LLM issues a query to either PubMed DB or NCT DB, its generated question is used as the search input, and relevant texts are retrieved via hybrid retrieval that combines BM25 \cite{robertson2009probabilistic} and embedding-based similarity. To mitigate the risk of label leakage, we apply a publication-date filter to ensure that all retrieved documents were publicly available prior to the start date of the clinical trial under consideration. This step is critical, as databases such as PubMed may include post hoc analyses or results of the trial itself, which—if accessed during feature construction—could leak outcome-related information and compromise the validity of the model. Similarly, any query made to the NCT DB excludes trials that began after the start date of the trial under consideration, ensuring that only information available at the time of the trial is used. 

\subsection{Feature Proposer}
\label{sec:feature_proposer}
The Feature Proposer\footnotemark{} module assumes the role of a biomedical expert or clinical researcher by generating conceptual feature ideas that are concretized in downstream stages. As shown in Figure~\ref{fig:framework}, we design two variants of this component: the Initializing Proposer, which operates during the first iteration, and the Iterative Proposer, which functions during the Monte Carlo Tree Search (MCTS) process.


\textbf{Initializing Proposer.}
The Initializing Proposer is designed to generate a foundational set of intuitive and expert-aligned features before any empirical evaluation occurs. It incorporates suggestions from two complementary sources:
\begin{itemize}[leftmargin=*]
\setlength\itemsep{-0.3em}
    \item Zero-Shot Proposer: This LLM is given only the task description and asked to suggest feature ideas based on its prior (parametric) knowledge, using chain-of-thought (CoT) reasoning. These features tend to be generic and broadly applicable.
    \item Factor-Based Proposer: This model receives individual labeled training samples (either positive or negative) and is tasked with identifying contributing factors to the observed label. It uses the ReAct framework and can query retrieval tools to perform more targeted and informed reasoning.
\end{itemize}

The combination of these two approaches balances general domain knowledge with sample-specific insights. To synthesize these ideas, CoT-enabled LLMs aggregate and summarize the outputs into a unified feature set, which is then passed to the Feature Planner for downstream use.

\textbf{Iterative Proposer.}
The Iterative Proposer differs from the Initializing Proposer in both input and output. It operates after each pipeline iteration, taking as input a recommendation from the Evaluator module. It then generates a single proposal of one of three types:
	1.	\texttt{Add} – introduces a new feature idea;
	2.	\texttt{Refine} – improves or augments an existing feature;
	3.	\texttt{Remove} – eliminates a feature deemed unhelpful or redundant.

\subsection{Feature Planner}
\label{sec:feature_planner}
The Feature Planner\footnotemark[\value{footnote}] is responsible for turning high-level feature ideas into structured, executable plans that downstream components can implement. Given a conceptual feature proposal from the Feature Proposer, this module outputs a detailed, step-by-step plan for feature construction, including an explicit schema that defines the format and expected output. This process relies on LLMs using chain-of-thought (CoT) reasoning and their demonstrated ability to perform complex task decomposition and procedural planning \cite{huang2024understanding}.

\subsection{Feature Builder}
\label{sec:feature_builder}
The Feature Builder\footnotemark[\value{footnote}] is responsible for assigning concrete values to each proposed feature for all clinical trials in the dataset, following the structured plans generated by the Feature Planner. 

During the initialization phase, the system processes a batch of feature ideas proposed by the Initializing Proposer. Since many of these features may be conceptually similar or share dependencies, we invoke the Feature Grouper to cluster them into logical groups. This grouping step ensures efficient use of the LLM’s context window and allows shared research steps to be reused across related features.

For each group, the system performs the following steps:

\begin{itemize}[leftmargin=*]
\setlength\itemsep{-0.3em}
    \item \textbf{Feature Researcher}: This module uses the ReAct paradigm \cite{yao2023react} to perform tool-augmented reasoning, retrieving information from external sources such as the PubMed DB or NCT DB. It enables the LLM to conduct targeted research, such as locating relevant trial histories, drug properties, or disease mechanisms, which are critical for building more complex and informative features.
    \item \textbf{Feature Builder}: After the necessary information is gathered, the Feature Builder executes the final step of information extraction \cite{xu2024large}. It uses CoT reasoning to convert retrieved evidence into structured values based on the schema defined in the feature plan. 
\end{itemize}

\subsection{Model Builder}
\label{sec:model_builder}
When the Feature Builder finishes running for all features for all the trials, the Model Builder, which is a function call, builds three classical machine learning models, Logistic Regression, Random Forest and XGBoost. 

\subsection{Evaluator}
\label{sec:evaluator}
The Evaluator\footnotemark[\value{footnote}] plays a critical role in guiding the Monte Carlo Tree Search (MCTS) process by generating  suggestions for improving model performance. These suggestions form the nodes explored in subsequent iterations of the \ourmethod\ pipeline. Following the design intuition behind the Initializing Proposer, we employ two complementary variants of the Evaluator:

\footnotetext{Prompts and example outputs are provided in Appendix~\ref{sec:appendix_prompt}.}

\begin{itemize}[leftmargin=*]
\setlength\itemsep{-0.3em}
    \item \textbf{Model-Based Evaluator}: This variant is provided with the model’s quantitative outputs, including the ROC-AUC score, feature importance, and the corresponding feature plans. Based on this information, the LLM is prompted with CoT to analyze which aspects of the current feature set may be lacking and to propose high-level improvement ideas. 
    \item \textbf{Error-Based Evaluator}: This variant receives all the information provided to the first version, but is additionally given a single misclassified validation example per iteration. Using ReAct reasoning, the LLM is employed to investigate potential reasons for the model's incorrect prediction. It retrieves relevant contextual information from external sources (e.g., PubMed DB, NCT DB) to understand what factors could have contributed to the correct classification and how those could be captured in a new or modified feature.
\end{itemize}
The suggestions generated by both Evaluators are aggregated to form a unified list of proposals. Each suggestion is treated as a potential next move in the MCTS algorithm, where it is passed to the Iterative Proposer.

\subsection{Monte Carlo Tree Search (MCTS)}
\label{sec:mcts}
We adopt Monte Carlo Tree Search (MCTS) \cite{coulom2006efficient} to efficiently explore the space of possible features within \ourmethod. Each node in the tree represents a distinct state of the feature set, and each edge corresponds to a transformation suggested by the Evaluator module—either an \texttt{Add}, \texttt{Refine}, or \texttt{Remove} action. At each iteration, \ourmethod\ selects the next action to apply by simulating multiple rollouts from the current state using the Upper Confidence Bound for Trees (UCT) criterion to balance exploration and exploitation:
\begin{equation}
  \label{eq:uct}
  \text{UCT}(x) = \frac{q(x)}{n(x)} + \alpha * \sqrt{\frac{\ln n(x_{parent})}{n(x)}},
\end{equation} where $q(x)$ is the cumulative reward, $n(x)$ is the number of times node $x$ has been visited, $n(x_{parent})$ is the visit count of the parent node, and $\alpha$ is the exploration weight that controls the trade-off between exploitation and exploration.

The resulting feature set is evaluated via model performance, and the Evaluator generates new improvement suggestions, which are then added as child nodes in the search tree. The MCTS process continues until the maximum number of iterations. At the end of the search, the system selects the feature set corresponding to the node with the best observed validation performance. 
\begin{table*}[ht!]
\small
\centering
\begin{minipage}{\linewidth}
\vspace{-0.5em}
\centering
\resizebox{\textwidth}{!}{
\begin{tabular}{lcccccccccccccc}
\toprule
\multirow{2}{*}{\textbf{Model}} & \multicolumn{3}{c}{\textbf{Phase I}} & \multicolumn{3}{c}{\textbf{Phase II}} & \multicolumn{3}{c}{\textbf{Phase III}}  \\
\cmidrule(lr){2-4} \cmidrule(lr){5-7} \cmidrule(lr){8-10} 
& PR-AUC & F1 & ROC-AUC & PR-AUC & F1 & ROC-AUC & PR-AUC & F1 & ROC-AUC \\
\midrule
LR       & 0.500 & 0.604 & 0.520 & 0.565 & 0.555 & 0.587 & 0.687 & 0.698 & 0.650  \\
RF           & 0.518 & 0.621 & 0.525 & 0.578 & 0.563 & 0.588 & 0.692 & 0.686 & 0.663 \\
XGBoost      & 0.513 & 0.621 & 0.518 & 0.586 & 0.570 & 0.600 & 0.697 & 0.696 & 0.667  \\
AdaBoost     & 0.519 & 0.622 & 0.526 & 0.586 & 0.583 & 0.603 & 0.701 & 0.695 & 0.670  \\
kNN + RF     & 0.531 & 0.625 & 0.538 & 0.594 & 0.590 & 0.597 & 0.707 & 0.698 & 0.678  \\
FFNN         & 0.547 & 0.634 & 0.550 & 0.604 & 0.599 & 0.611 & 0.747 & 0.748 & 0.681  \\
DeepEnroll   & 0.568 & 0.648 & 0.575 & 0.600 & 0.598 & 0.625 & 0.777 & 0.786 & 0.699 \\
COMPOSE     & 0.564 & 0.658 & 0.571 & 0.604 & 0.597 & 0.628 & 0.782 & 0.792 & 0.700  \\
HINT       & 0.567 & 0.665 & 0.576 & 0.629 & 0.620 & 0.645 & 0.811 & 0.847 & 0.723 \\
SPOT       & 0.689 & 0.714 & 0.660 & 0.685 & 0.656 & 0.630 & 0.856 & 0.857 & 0.711  \\
MMFusion  & 0.579 & 0.701 & 0.782 & 0.510 & 0.590 & 0.771 & 0.638 & 0.742 & 0.741  \\
\textbf{\ourmethod}  & \textbf{0.710} & \textbf{0.595} & \textbf{0.753} & \textbf{0.512} & \textbf{0.386} & \textbf{0.639} & \textbf{0.697} & \textbf{0.760} & \textbf{0.702}  \\
\bottomrule
\end{tabular}
}
\caption{Model performance on clinical trial outcome prediction. \ourmethod\ numbers represent its test performance on a subset of 100 samples using the best model selected on the validation set.}
\vspace{-1em}
\label{tab:full_results}
\end{minipage}
\end{table*}
\section{Experiments}
\subsection{Experimental Setup}
The algorithm for the \ourmethod\ framework is provided in Figure~\ref{fig:alg}. We use \texttt{gpt-4o-mini}\footnote{\url{https://openai.com/index/gpt-4o-mini-advancing-cost-efficient-intelligence/}} as the backbone LLM, with a temperature of 0. For the Initializing Proposer, we randomly select 3 positive and 3 negative samples from the training set and process each through the Factor-Based Proposer. Similarly, for the Evaluator module, 3 misclassified validation samples are randomly selected and evaluated using the Error-Based Evaluator. As a result, each node in the Monte Carlo Tree Search (MCTS) generates a maximum of 6 potential improvement proposals.

For the MCTS process, we set the exploration weight $\alpha$ in the UCT formula to 1.0 to bias the search toward deeper, potentially more promising parts of the tree. We use ROC-AUC as the primary performance metric to guide the tree search, in alignment with prior work such as \citet{lo2019machine,siah2021predicting,fu2022hint,chen2024trialbench}. We limit the number of rollouts to 10 and the maximum depth to 10. Each MCTS run on a subset of 100 training and 100 validation samples incurs an approximate cost of \$150, with the feature building process responsible for the majority of the compute cost due to in-depth retrieval and reasoning.

\subsection{Dataset}
We evaluate \ourmethod\ primarily on the Trial Approval Prediction task from the TrialBench benchmark~\cite{chen2024trialbench}, which includes 24,468 training and 6,215 test samples. To ensure a fair comparison with prior benchmarks and to remain within our computational budget, we perform stratified sampling based on the label distribution to select 100 training and 100 validation samples from the original training set, along with 100 samples from the original test set.

To demonstrate the generalizability of \ourmethod\ across different clinical prediction tasks, we also evaluate it on three additional datasets from \citet{chen2024trialbench}: Patient Dropout, Mortality and Adverse Event Prediction. For these datasets, we adopt the same stratified sampling strategy as in the trial approval task. To manage computational cost, we restrict evaluation to the subset of phase I trials and limit the MCTS process to a maximum of 5 rollouts per run, terminating early once reasonable predictive performance is achieved.

Although the benchmark datasets come with a base set of features including molecule SMILES and disease codes, for this study the agent is provided with only the unique trial identifiers (NCT IDs), with which it derives the set of features it believes to be important for each task.

\subsection{Baseline Methods}
For the trial approval prediction task, we compare \ourmethod\ with a range of baselines including both traditional machine learning models and state-of-the-art deep learning approaches. Following the setup in \citet{fu2022hint}, we include Logistic Regression (LR) \cite{lo2019machine, siah2021predicting}, Random Forest (RF) \cite{lo2019machine, siah2021predicting}, XGBoost \cite{rajpurkar2020evaluation,siah2021predicting}, Adaptive Boosting (AdaBoost) \cite{fan2020application}, k-Nearest Neighbors (kNN) combined with RF \cite{lo2019machine}, Feedforward Neural Networks (FFNN) \cite{tranchevent2019deep}, DeepEnroll \cite{zhang2020deepenroll}, COMPOSE \cite{gao2020compose}, and HINT \cite{fu2022hint}. 

In addition, we incorporate the results from SPOT \cite{wang2023spot} and the deep learning method ("MMFusion") introduced in TrialBench \cite{chen2024trialbench} as additional benchmarks. The overall comparison is presented in Table~\ref{tab:full_results}. For a fair evaluation, we report the test performance of \ourmethod\ using the best model selected on the validation set during the MCTS process.

It is important to note that all baselines reported by \citet{fu2022hint} are trained and evaluated on the TOP dataset. In contrast, the benchmark results from TrialBench and our method use a more recent version of this dataset released by \citet{chen2024trialbench}. 

\begin{table}[t]
\centering
\small
\setlength{\tabcolsep}{4pt}
\begin{tabular}{@{}llccc@{}}
\toprule
\textbf{Task} & \textbf{Method} & \textbf{PR-AUC} & \textbf{F1} & \textbf{ROC-AUC} \\
\midrule
\multirow{2}{*}{Patient Dropout} 
    & MMFusion & 0.691 & 0.718 & 0.723 \\
    & \ourmethod     & \textbf{0.795} & \textbf{0.718} & \textbf{0.711} \\
\midrule
\multirow{2}{*}{Mortality} 
    & MMFusion & 0.610 & 0.745 & 0.900 \\
    & \ourmethod     & \textbf{0.560} & \textbf{0.732} & \textbf{0.852} \\
\midrule
\multirow{2}{*}{Adverse Event} 
    & MMFusion & 0.726 & 0.793 & 0.874 \\
    & \ourmethod     & \textbf{0.796} & \textbf{0.731} & \textbf{0.831} \\
\bottomrule
\vspace{-1em}
\end{tabular}
\caption{Model performance on predicting patient dropout, mortality, and adverse events for phase I trials. \ourmethod\ was configured with a maximum rollout of 5 and a search depth of 10. \ourmethod\
numbers represent the test performance on a subset of 100 samples using the best model selected on the validation
set.}
\label{tab:other_trialbench_results}
\end{table}

\subsection{Results}
\subsubsection{\ourmethod\ on Trial Approval Prediction} 
The performance of \ourmethod\ is summarized in Table~\ref{tab:full_results}. With a maximum of 10 rollouts and a tree depth limit of 10, \ourmethod\ achieves ROC-AUC scores of 0.753, 0.639, and 0.702 on the test set for Phase I, II, and III, respectively. These results are comparable to existing benchmarks, particularly recent deep learning approaches such as HINT, SPOT and TrialBench.

\subsubsection{Impact of Maximum Rollouts} 

\begin{figure}[t]
    \centering
    \includegraphics[width=\linewidth]{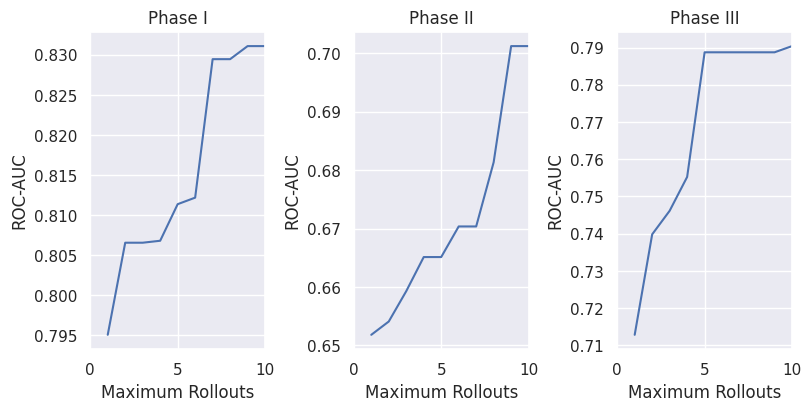}
    \caption{Average test set ROC-AUC of the top 5 models under varying maximum rollout limits in MCTS. Models are ranked by test set performance to smooth out noise and illustrate overall trends.}
    \vspace{-1em}
    \label{fig:max_rollouts}
\end{figure}

To assess the effect of increasing the maximum number of rollouts in MCTS, we track test set performance across different rollout limits for all three phases, as shown in Figure~\ref{fig:max_rollouts}. Given the small sample sizes, we report the average ROC-AUC of the top 5 models (ranked by test set performance) to smooth out noise and better capture the performance trend. While some fluctuations in test scores remain—likely due to overfitting on the validation set or distributional shifts between validation and test subsets—the overall trend indicates improved test performance with increased rollouts. This suggests that more exploration in the MCTS tree enables the discovery of more informative features and strategies.

However, we do observe plateauing of the performance across the rollouts. A plausible explanation is the limited dataset size: with only 100 training and 100 validation samples, the model may lack sufficient data to capture the diverse and complex relationships necessary for generalization. For instance, trials involving different disease categories may require distinct sets of features, which are difficult to learn with such small sample sizes. Nevertheless, \ourmethod\ demonstrates strong and consistent performance despite these constraints, which highlights its potential for efficient and interpretable feature discovery in low-data regimes.

\subsubsection{\ourmethod\ on Other Clinical Tasks}

We evaluate the performance of \ourmethod\ on three additional clinical prediction tasks from \citet{chen2024trialbench}: Patient Dropout, Mortality, and Adverse Event prediction. The results are summarized in Table~\ref{tab:other_trialbench_results}. With a maximum of 5 MCTS rollouts, \ourmethod\ achieves ROC-AUC scores of 0.711, 0.852, and 0.831 on the respective tasks. 

Due to the recency of these datasets, published baselines are limited, and we compare primarily against the MMFusion results. The comparable performance between \ourmethod\ and MMFusion across these diverse prediction tasks demonstrates the robustness and generalizability of our method beyond the trial approval task.

\subsection{Case Study}
\label{sec:case_study}

\subsubsection{Specific Trial Outcome Prediction}

The use of classical machine learning models in \ourmethod\ allows us to leverage interpretability techniques such as SHAP (SHapley Additive exPlanations)~\cite{lundberg2017unified} to understand model predictions. We present selected correctly classified test examples and their associated SHAP value analyses, which highlight the impact of individual features on the predicted outcome. Full SHAP plots for these examples are included in the Appendix~\ref{sec:appendix_shap}.

\begin{itemize}[leftmargin=*]
\setlength\itemsep{-0.3em}
    \item Phase III study on PEG-Intron in HIV-infected Patients by Merck Sharp \& Dohme LLC (NCT ID: \texttt{NCT00035360}): The model outputs a low approval probability of 0.244. SHAP analysis indicates that the primary outcome measure and duration of treatment were the most influential features contributing to the negative prediction.
    
    \item Phase III trial on AAT-023 (Zuragen) vs. Heparin in Catheter-Related Bloodstream Infections by Ash Access Technology (NCT ID: \texttt{NCT00628680}): The model outputs a high approval probability of 0.895. Feature contributions from treatment duration and inclusion criteria count were key to the positive prediction.
    
    \item Phase I study on Birabresib in Selected Advanced Solid Tumors by Merck Sharp \& Dohme LLC (NCT ID: \texttt{NCT02698176}): The model assigns a low approval probability of 0.197. SHAP values suggest that the negative outcome was driven by factors such as trial design, geographical location, route of administration, and strictness of eligibility criteria.
\end{itemize}

\subsubsection{Evolution of Feature Set} 

We trace one of the search paths in the MCTS for Phase I trial outcome prediction to better understand how MCTS iteratively improves and refines the feature set. Beginning with an initial feature set suggested by the Initializing Feature Proposer, one of the explored paths incorporated recommendations from the Evaluator, including:

\begin{itemize}[leftmargin=*]
\setlength\itemsep{-0.3em}
    \item \texttt{Add} \textit{adverse event rate}, to capture the frequency of adverse events in previous trials within the same therapeutic area. This suggestion originated from the Model-Based Evaluator, which recommended to build this feature using data from ClinicalTrials.gov and PubMed literature.
    
    \item \texttt{Refine} \textit{trial design elements}, to explicitly include key design attributes such as randomization, blinding, and control groups. This recommendation came from the Example-Based Evaluator, which referenced a trial with a quadruple-masking design—an aspect that would have improved the reliability of trial results.
    
    \item \texttt{Remove} \textit{intervention type}, due to its low feature importance in the model output. Although this feature was part of the initial proposal, the Model-Based Evaluator determined it to be unhelpful, as most trials in the curated dataset are drug trials~\cite{chen2024trialbench}.
\end{itemize}

\section{Conclusion}
Our proposed \ourmethod\ framework integrates large language models with classical machine learning to automate clinical trial prediction using only trial identifiers and outcome labels. Inspired by how biomedical experts approach data-driven tasks, \ourmethod\ leverages LLMs not to give direct probability estimations, but to propose, plan, and construct meaningful features through research and reasoning. Our use of Monte Carlo Tree Search (MCTS) allows for iterative refinement guided by performance feedback and error analysis, which significantly reduces the need for human intervention. Experiments on the TrialBench dataset demonstrate that \ourmethod\ achieves competitive performance compared to strong baselines and expert-tuned models. Beyond its promising results, \ourmethod\ can be applied in high-stakes domains where interpretability remains a major challenge.  

\section*{Limitations}
This work has several limitations that point to potential directions for future improvement. First, the retrieval component of our system is currently limited to two data sources—ClinicalTrials.gov (NCT) and PubMed—due to concerns around knowledge cutoff and the risk of incorporating information not available at the time of the trial. While this conservative design choice ensures label integrity, it may restrict the richness and variety of extracted features. Second, we intentionally do not perform hyperparameter optimization, as the focus of this work is to provide a proof-of-concept rather than to achieve maximum predictive performance. Nonetheless, we believe that the framework could benefit from more rollouts, larger LLMs, and systematic tuning to improve overall accuracy. Finally, the Monte Carlo Tree Search (MCTS) is currently constrained to exploring the space of feature proposals. Ideally, the Evaluator module could be made more intelligent by identifying and attributing underperformance to specific pipeline components, such as differentiating between flawed reasoning by the Feature Researcher versus improper execution by the Feature Builder, which would enable more targeted improvements.


\bibliography{anthology,custom}

\clearpage
\appendix
\section{Example SHAP Plots}
\label{sec:appendix_shap}

\begin{figure}[h]
    \centering
    \includegraphics[width=2.0\columnwidth]{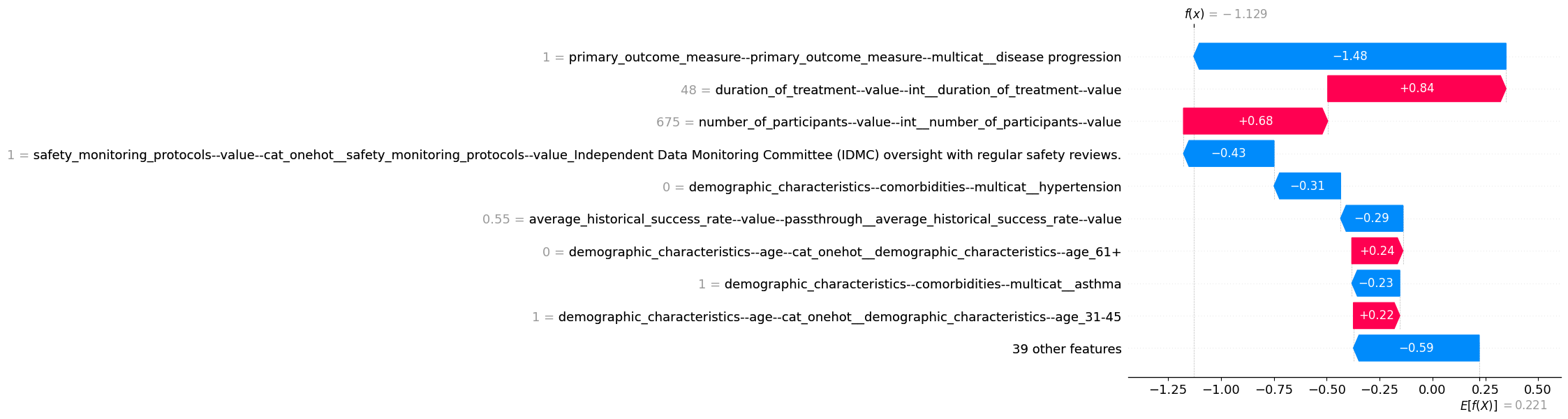}
    \caption{SHAP Values for NCT ID: \texttt{NCT00035360}}
    \label{fig:hiv}
\end{figure}

\begin{figure}[h]
    \centering
    \includegraphics[width=2.0\columnwidth]{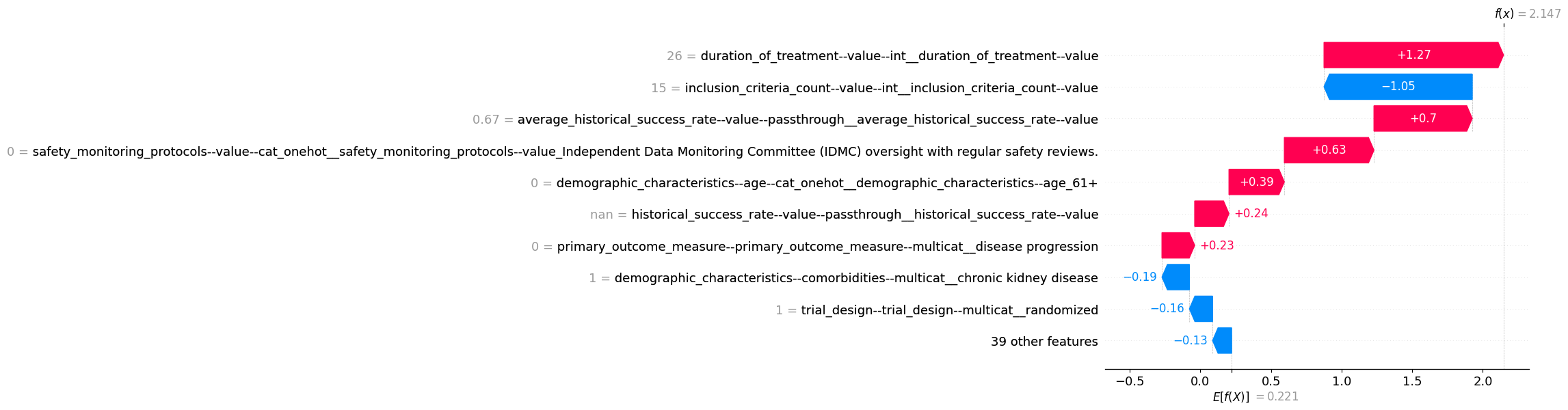}
    \caption{SHAP Values for NCT ID: \texttt{NCT00628680}}
    \label{fig:zuragen}
\end{figure}

\begin{figure}[h]
    \centering
    \includegraphics[width=2.0\columnwidth]{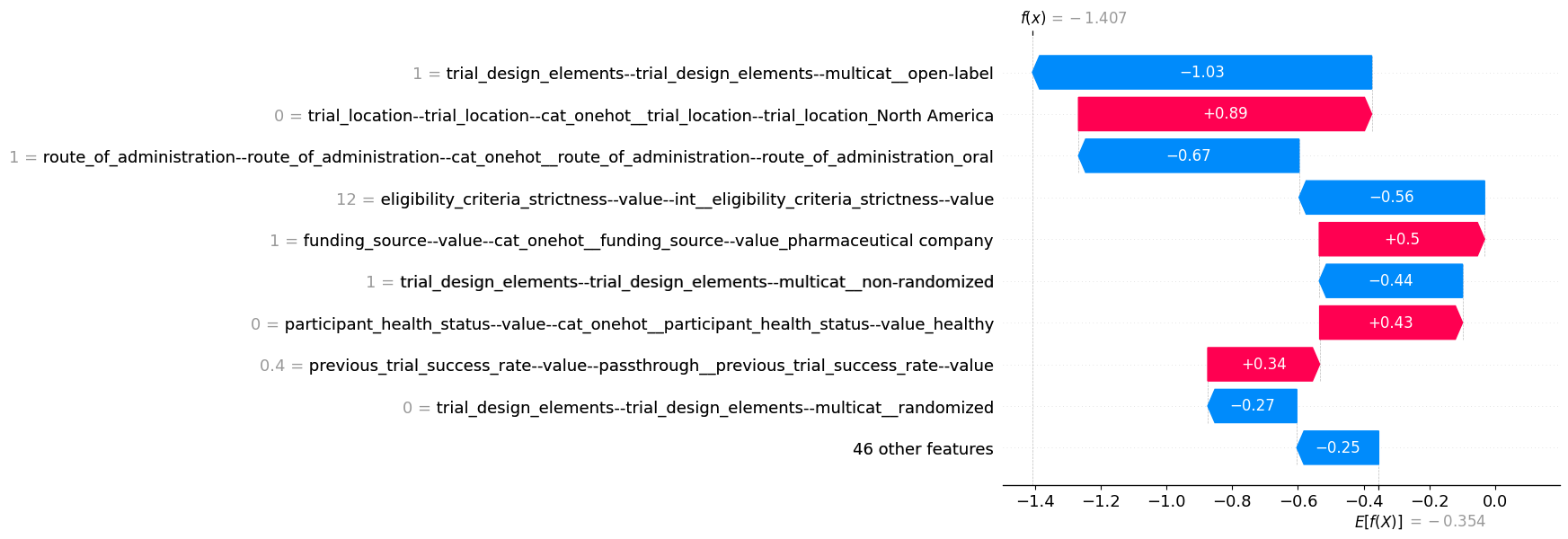}
    \caption{SHAP Values for NCT ID: \texttt{NCT02698176}}
    \label{fig:birab}
\end{figure}

\clearpage
\section{Example LLM Interaction}
\label{sec:appendix_prompt}

We present the prompts and example outputs of the two variants of the Initializing Feature Proposer in Figures~\ref{fig:zeroshot_proposer} and \ref{fig:factor_based_proposer}. The Zero-shot Proposer draws on its prior knowledge of clinical trial prediction tasks, and suggests generic features such as intervention type, number of participants, research team experience, and the historical success rate of similar trials. While direct generation using LLMs is effective, it is often not sufficiently comprehensive~\cite{feng2024bird}.

In contrast, the Factor-Based Proposer takes an example trial as input and leverages local context to generate more fine-grained feature ideas, such as participant health status and safety profiles. However, the provided context can bias the LLM’s output, resulting in a narrower scope. For instance, one of the proposed features in Figure~\ref{fig:factor_based_proposer} refers to previous Flavivirus exposure, which is highly specific to the given trial example.

To balance generality and specificity, we employ an LLM summarizer to synthesize outputs from both Proposers. This final step filters out overly specific ideas and ensures that the proposed features are generalizable across a wide range of clinical trials.

\lstdefinestyle{json}{
  basicstyle=\ttfamily\scriptsize,
  breaklines=true,
  showstringspaces=false,
  columns=fullflexible,
  frame=single
}

\begin{figure*}[hbt!]
\small
\begin{tcolorbox}[title=Zero-shot Feature Proposer Prompt Instructions, colback=gray!5, colframe=black, fonttitle=\bfseries]

\textbf{Role:} You are an experienced clinical researcher skilled at proposing features for a machine learning model.

\textbf{Task:} Propose a comprehensive list of feature ideas (at least 10) for this model. Be as exhaustive and as detailed as possible in describing the feature.

\textbf{Data Sources:} The features should be built off data from ClinicalTrials.gov and from past scholarly research from PubMed.

\textbf{Constraints:}
\begin{itemize}
    \item Feature type must be one of: integer, float, boolean, categorical, or multicategorical.
    \item Features should not be a composite of multiple factors.
    \item Features should not be the output of another machine learning model.
    \item Features should only require data retrievable from ClinicalTrials.gov or PubMed.
\end{itemize}

\textbf{Input:} Predict the outcome of a phase 1 clinical trial (1 = success, 0 = failure) at the beginning stages of a trial.

\end{tcolorbox}

\vspace{1em}

\begin{tcolorbox}[title=Example LLM Response, colback=gray!3, colframe=black, fonttitle=\bfseries]
\begin{lstlisting}[style=json]
[
  {
    "feature_name": "intervention_type",
    "description": "Categorical feature indicating the type of intervention (e.g., drug, device, behavioral)."
  },
  {
    "feature_name": "number_of_participants",
    "description": "Integer feature representing the total number of participants enrolled in the trial."
  },
  {
    "feature_name": "age_range",
    "description": "Categorical feature indicating the age range of participants (e.g., 18-30, 31-50, 51+)."
  },
  {
    "feature_name": "gender_inclusion",
    "description": "Boolean feature indicating whether both genders are included in the trial."
  },
  {
    "feature_name": "previous_trial_success_rate",
    "description": "Float feature representing the historical success rate of similar trials in the same therapeutic area."
  },
  {
    "feature_name": "research_team_experience",
    "description": "Integer feature quantifying the number of years of experience of the principal investigator in conducting clinical trials."
  },
  {
    "feature_name": "funding_source",
    "description": "Categorical feature indicating the source of funding (e.g., government, pharmaceutical company, non-profit)."
  },
  {
    "feature_name": "primary_outcome_measure",
    "description": "Categorical feature describing the primary outcome measure (e.g., safety, efficacy, pharmacokinetics)."
  },
  {
    "feature_name": "trial_location",
    "description": "Categorical feature indicating the geographical location of the trial (e.g., North America, Europe, Asia)."
  },
  {
    "feature_name": "eligibility_criteria_strictness",
    "description": "Integer feature representing the number of strict eligibility criteria defined for participant selection."
  }
]
\end{lstlisting}
\end{tcolorbox}

\caption{Example prompt and response for the Zero-shot Feature Proposer agent.}
\label{fig:zeroshot_proposer}
\end{figure*}

\begin{figure*}[hbt!]
\small
\begin{tcolorbox}[title=Factor-Based Feature Proposer Prompt Instructions, colback=gray!5, colframe=black, fonttitle=\bfseries]

\textbf{Role:} You are an experienced clinical researcher.

\textbf{Task:} Analyze clinical trials to deduce key factors that contributed to a particular trial outcome (success or failure), in order to inform feature design for a machine learning model.

\textbf{Instructions:}
\begin{itemize}
    \item Provide at least 5 concise, generalizable factors relevant to future trials.
    \item Factors can be drawn from trial context, historical clinical trials, or PubMed literature.
\end{itemize}

\textbf{Input:} \texttt{NCTID: NCT01224639 (success)}

\end{tcolorbox}

\vspace{1em}

\begin{tcolorbox}[title=Example LLM Response, colback=gray!5, colframe=black, fonttitle=\bfseries]
\begin{lstlisting}[style=json]
[
  {
    "name": "Route of Administration",
    "description": "The method of vaccine delivery (subcutaneous vs. intradermal) can significantly affect the immune response and safety profile."
  },
  {
    "name": "Dosing Regimen",
    "description": "The amount of vaccine administered (low vs. high dose) influences the immunogenicity and reactogenicity, impacting overall trial outcomes."
  },
  {
    "name": "Previous Flavivirus Exposure",
    "description": "Participants' prior exposure to flavivirus can alter their immune response to the vaccine, affecting safety and efficacy."
  },
  {
    "name": "Safety and Reactogenicity Profiles",
    "description": "Historical data on adverse events and tolerability from similar trials can guide expectations for safety in the current trial."
  },
  {
    "name": "Participant Health Status",
    "description": "The overall health and eligibility criteria of participants, including age and pre-existing conditions, can influence trial outcomes."
  }
]
\end{lstlisting}
\end{tcolorbox}

\caption{Example prompt and response for the Factor-Based Feature Proposer agent.}
\label{fig:factor_based_proposer}
\end{figure*}

\begin{figure*}[hbt!]
\small
\begin{tcolorbox}[title=Feature Planner Prompt Instructions, colback=gray!5, colframe=black, fonttitle=\bfseries]

\textbf{Role:} You are an expert data scientist.

\textbf{Task:} You are given an idea for a single feature to be used in a machine learning model for a clinical trial task. For this single feature, you are defining a feature schema for your co-workers to construct the feature for each clinical trial.

\textbf{Instructions:}
The final built feature should be a JSON object
\begin{itemize}
    \item If there's only a single value, it should be a JSON with a single key "value" and the value.
    \item If there are multiple values, it should be a JSON with multiple keys, each key corresponding to a sub-feature name, and the value corresponding to the sub-feature value.
\end{itemize}
The schema and instruction should be as simple as possible to represent the feature idea.

Your instruction should be clear, and allow for the feature to be computed consistently and reliably. The instruction needs to be explicit and avoid ambiguity since multiple teams are working together. For e.g., if weights need to be assigned, they should be explicitly defined in the instructions. The feature should be built off data from ClinicalTrials.gov and from past scholarly research from PubMed.

\textbf{Input:} Idea: Categorical feature indicating the geographical location of the trial (e.g., North America, Europe, Asia).

\end{tcolorbox}

\vspace{1em}

\begin{tcolorbox}[title=Example LLM Response (Phase I Plans), colback=gray!5, colframe=black, fonttitle=\bfseries]
\begin{lstlisting}[style=json]
{
  "intervention_type": {
    "feature_name": "intervention_type",
    "feature_idea": "Categorical feature indicating the type of intervention (e.g., drug, device, behavioral).",
    "feature_type": {
      "value": "categorical"
    },
    "data_sources": [
      "pubmed",
      "current_trial_summary",
      "related_clinical_trials"
    ],
    "example_values": [
      {
        "value": "drug"
      },
      {
        "value": "device"
      },
      {
        "value": "behavioral"
      }
    ],
    "possible_values": {
      "value": [
        "drug",
        "device",
        "behavioral"
      ]
    },
    "feature_instructions": "Research and compile a list of intervention types used in clinical trials. The feature should categorize interventions into distinct types such as \"drug\", \"device\", and \"behavioral\". Ensure that the categorization is consistent with definitions used in clinical trial registries and literature. Use data from ClinicalTrials.gov and PubMed to validate the types of interventions. The final output should be a JSON object with a single key \"value\" representing the intervention type."
  },
    "number_of_participants": {
    "feature_name": "number_of_participants",
    "feature_idea": "Integer feature representing the total number of participants enrolled in the trial.",
    "feature_type": {
      "value": "integer"
    },
    "data_sources": [
      "current_trial_summary",
      "related_clinical_trials"
    ],
\end{lstlisting}
\end{tcolorbox}
\end{figure*}

\begin{figure*}[hbt!]
\small
\begin{tcolorbox}[title=Example LLM Response (Phase I Plans) (cont'd), colback=gray!5, colframe=black, fonttitle=\bfseries]
\begin{lstlisting}[style=json]
    "example_values": [
      {
        "value": 50
      },
      {
        "value": 100
      },
      {
        "value": 200
      }
    ],
    "possible_values": {},
    "feature_instructions": "To construct the feature \"number_of_participants,\" extract the total number of participants enrolled in the clinical trial from the current trial summary data. Ensure that the value is recorded as an integer. If the data is not available, it should be marked as null or zero. This feature should be updated regularly as participant enrollment progresses throughout the trial."
  },
  "age_range": {
    "feature_name": "age_range",
    "feature_idea": "Categorical feature indicating the age range of participants (e.g., 18-30, 31-50, 51+).",
    "feature_type": {
      "age_range": "categorical"
    },
    "data_sources": [
      "pubmed",
      "current_trial_summary",
      "related_clinical_trials"
    ],
    "example_values": [
      {
        "age_range": "18-30"
      },
      {
        "age_range": "31-50"
      },
      {
        "age_range": "51+"
      }
    ],
    "possible_values": {
      "age_range": [
        "18-30",
        "31-50",
        "51+"
      ]
    },
    "feature_instructions": "Research and compile the age ranges of participants in the clinical trials. The age ranges should be categorized as follows: 18-30, 31-50, and 51+. Ensure that the data is sourced from ClinicalTrials.gov and relevant literature from PubMed. Each trial should have a clearly defined age range based on the participants enrolled. If a trial includes participants from multiple age ranges, select the range that represents the majority of participants."
  },
    "gender_inclusion": {
    "feature_name": "gender_inclusion",
    "feature_idea": "Boolean feature indicating whether both genders are included in the trial.",
    "feature_type": {
      "value": "boolean"
    },
    "data_sources": [
      "current_trial_summary",
      "related_clinical_trials"
    ],
    "example_values": [
      {
        "value": true
      },
      {
        "value": false
      }
    ],
    "possible_values": {},
    "feature_instructions": "Research the clinical trial protocols to determine if both genders are included in the trial. This should be assessed based on the eligibility criteria listed in the trial summary. If both genders are included, set the value to true; otherwise, set it to false. Ensure that the assessment is consistent across all trials by strictly adhering to the eligibility criteria provided in the trial documentation."
  },
\end{lstlisting}
\end{tcolorbox}
\end{figure*}

\begin{figure*}[hbt!]
\small
\begin{tcolorbox}[title=Example LLM Response (Phase I Plans) (cont'd), colback=gray!5, colframe=black, fonttitle=\bfseries]
\begin{lstlisting}[style=json]
  "previous_trial_success_rate": {
    "feature_name": "previous_trial_success_rate",
    "feature_idea": "Float feature representing the historical success rate of similar trials in the same therapeutic area.",
    "feature_type": {
      "value": "float"
    },
    "data_sources": [
      "pubmed",
      "related_clinical_trials",
      "current_trial_summary"
    ],
    "example_values": [
      {
        "value": 0.75
      },
      {
        "value": 0.6
      },
      {
        "value": 0.85
      }
    ],
    "possible_values": {},
    "feature_instructions": "To compute the previous trial success rate, gather data from clinical trials in the same therapeutic area. Focus on phase 1 trials and extract the outcomes (success or failure) of these trials. Calculate the success rate by dividing the number of successful trials by the total number of trials analyzed. The resulting value should be a float representing the success rate as a percentage (e.g., 0.75 for 75% success). Ensure that the data is up-to-date and relevant to the current therapeutic area being studied."
  },
  "research_team_experience": {
    "feature_name": "research_team_experience",
    "feature_idea": "Integer feature quantifying the number of years of experience of the principal investigator in conducting clinical trials.",
    "feature_type": {
      "value": "integer"
    },
    "data_sources": [
      "pubmed",
      "current_trial_summary"
    ],
    "example_values": [
      {
        "value": 10
      },
      {
        "value": 5
      },
      {
        "value": 15
      }
    ],
    "possible_values": {},
    "feature_instructions": "To construct the feature \"research_team_experience\", gather data on the principal investigator's years of experience in conducting clinical trials. This information can typically be found in the principal investigator's profile on ClinicalTrials.gov or through their published research on PubMed. Ensure that the experience is quantified in whole years and is accurately recorded. If the PI has experience in multiple trials, sum the years of experience across all relevant trials to provide a total. This feature should be represented as a single integer value in the final JSON output."
  },
  "funding_source": {
    "feature_name": "funding_source",
    "feature_idea": "Categorical feature indicating the source of funding (e.g., government, pharmaceutical company, non-profit).",
    "feature_type": {
      "value": "categorical"
    },
    "data_sources": [
      "current_trial_summary",
      "pubmed"
    ],
    "example_values": [
      {
        "value": "government"
      },
      {
        "value": "pharmaceutical company"
      },
      {
        "value": "non-profit"
      }
    ],
\end{lstlisting}
\end{tcolorbox}
\end{figure*}

\begin{figure*}[hbt!]
\small
\begin{tcolorbox}[title=Example LLM Response (Phase I Plans) (cont'd), colback=gray!5, colframe=black, fonttitle=\bfseries]
\begin{lstlisting}[style=json]
    "possible_values": {
      "value": [
        "government",
        "pharmaceutical company",
        "non-profit"
      ]
    },
    "feature_instructions": "Research and compile data on the funding sources for each clinical trial. The funding source should be categorized into three main types: government, pharmaceutical company, and non-profit. Ensure that the data is sourced from ClinicalTrials.gov and relevant literature from PubMed. The final output should be a JSON object with a single key \"value\" representing the funding source."
  },
  "primary_outcome_measure": {
    "feature_name": "primary_outcome_measure",
    "feature_idea": "Categorical feature describing the primary outcome measure (e.g., safety, efficacy, pharmacokinetics).",
    "feature_type": {
      "value": "categorical"
    },
    "data_sources": [
      "pubmed",
      "current_trial_summary",
      "related_clinical_trials"
    ],
    "example_values": [
      {
        "value": "safety"
      },
      {
        "value": "efficacy"
      },
      {
        "value": "pharmacokinetics"
      }
    ],
    "possible_values": {
      "value": [
        "safety",
        "efficacy",
        "pharmacokinetics",
        "tolerability",
        "biomarkers"
      ]
    },
    "feature_instructions": "Research and compile the primary outcome measures from clinical trials listed on ClinicalTrials.gov and relevant literature from PubMed. Ensure to categorize the outcome measures into predefined categories such as safety, efficacy, pharmacokinetics, tolerability, and biomarkers. The feature should be represented as a JSON object with a single key \"value\" that holds the categorical outcome measure. Consistency in categorization is key, so refer to existing literature for definitions and examples of each category."
  },
  "trial_location": {
    "feature_name": "trial_location",
    "feature_idea": "Categorical feature indicating the geographical location of the trial (e.g., North America, Europe, Asia).",
    "feature_type": {
      "trial_location": "categorical"
    },
    "data_sources": [
      "current_trial_summary",
      "related_clinical_trials"
    ],
    "example_values": [
      {
        "trial_location": "North America"
      },
      {
        "trial_location": "Europe"
      },
      {
        "trial_location": "Asia"
      }
    ],
    "possible_values": {
      "trial_location": [
        "North America",
        "Europe",
        "Asia",
        "South America",
        "Africa",
        "Oceania"
      ]
    },
\end{lstlisting}
\end{tcolorbox}
\end{figure*}

\begin{figure*}[hbt!]
\small
\begin{tcolorbox}[title=Example LLM Response (Phase I Plans) (cont'd), colback=gray!5, colframe=black, fonttitle=\bfseries]
\begin{lstlisting}[style=json]
    "feature_instructions": "Research and compile the geographical locations of clinical trials from ClinicalTrials.gov. Classify each trial's location into one of the predefined categories: North America, Europe, Asia, South America, Africa, or Oceania. Ensure that the categorization is consistent and based on the trial's registered location. This feature should be extracted from the trial's summary data and should be formatted as a categorical variable in the final JSON output."
  },
  "eligibility_criteria_strictness": {
    "feature_name": "eligibility_criteria_strictness",
    "feature_idea": "Integer feature representing the number of strict eligibility criteria defined for participant selection.",
    "feature_type": {
      "value": "integer"
    },
    "data_sources": [
      "current_trial_summary",
      "pubmed"
    ],
    "example_values": [
      {
        "value": 5
      },
      {
        "value": 10
      },
      {
        "value": 3
      }
    ],
    "possible_values": {},
    "feature_instructions": "To construct the feature \"eligibility_criteria_strictness\", review the eligibility criteria listed in the trial's summary on ClinicalTrials.gov. Count the number of strict criteria that are defined for participant selection. Each criterion should be considered strict if it imposes specific limitations on who can participate in the trial (e.g., age limits, health conditions, prior treatments). Record this count as an integer value. Ensure consistency in how criteria are evaluated to maintain reliability across different trials."
  },
  "route_of_administration": {
    "feature_name": "route_of_administration",
    "feature_idea": "Categorical feature indicating the method of intervention delivery (e.g., oral, intravenous).",
    "feature_type": {
      "route_of_administration": "categorical"
    },
    "data_sources": [
      "pubmed",
      "current_trial_summary",
      "related_clinical_trials"
    ],
    "example_values": [
      {
        "route_of_administration": "oral"
      },
      {
        "route_of_administration": "intravenous"
      },
      {
        "route_of_administration": "subcutaneous"
      }
    ],
    "possible_values": {
      "route_of_administration": [
        "oral",
        "intravenous",
        "subcutaneous",
        "intramuscular",
        "topical",
        "inhalation"
      ]
    },
    "feature_instructions": "Research and compile data on the various routes of administration used in clinical trials. Ensure to categorize each method of intervention delivery accurately. The feature should be built from data available on ClinicalTrials.gov and relevant scholarly articles from PubMed. Each trial should be assessed for its route of administration, and the corresponding value should be recorded in the feature schema. The possible values should include all common routes of administration used in clinical trials."
  },
\end{lstlisting}
\end{tcolorbox}
\end{figure*}

\begin{figure*}[hbt!]
\small
\begin{tcolorbox}[title=Example LLM Response (Phase I Plans) (cont'd), colback=gray!5, colframe=black, fonttitle=\bfseries]
\begin{lstlisting}[style=json]
  "dosing_regimen": {
    "feature_name": "dosing_regimen",
    "feature_idea": "Categorical feature describing the dosing schedule (e.g., single dose, multiple doses).",
    "feature_type": {
      "dosing_regimen": "categorical"
    },
    "data_sources": [
      "current_trial_summary",
      "pubmed"
    ],
    "example_values": [
      {
        "dosing_regimen": "single dose"
      },
      {
        "dosing_regimen": "multiple doses"
      }
    ],
    "possible_values": {
      "dosing_regimen": [
        "single dose",
        "multiple doses",
        "loading dose",
        "maintenance dose",
        "intermittent dosing"
      ]
    },
    "feature_instructions": "Research and compile the dosing regimens used in phase 1 clinical trials. Classify the regimens into categories such as \"single dose\", \"multiple doses\", \"loading dose\", \"maintenance dose\", and \"intermittent dosing\". Ensure that the categorization is consistent across trials and that each trial's dosing regimen is accurately represented. Use data from ClinicalTrials.gov and relevant literature from PubMed to validate the dosing schedules."
  },
  "participant_health_status": {
    "feature_name": "participant_health_status",
    "feature_idea": "Categorical feature indicating the overall health status of participants (e.g., healthy, comorbid conditions).",
    "feature_type": {
      "value": "categorical"
    },
    "data_sources": [
      "pubmed",
      "current_trial_summary",
      "related_clinical_trials"
    ],
    "example_values": [
      {
        "value": "healthy"
      },
      {
        "value": "comorbid conditions"
      }
    ],
    "possible_values": {
      "value": [
        "healthy",
        "comorbid conditions",
        "chronic illness",
        "acute condition",
        "unknown"
      ]
    },
    "feature_instructions": "Research and compile data on the health status of participants in phase 1 clinical trials. The health status should be categorized into predefined groups such as \"healthy\", \"comorbid conditions\", \"chronic illness\", \"acute condition\", and \"unknown\". Ensure that the categorization is consistent across trials and based on the information available in clinical trial summaries and relevant literature. Document the health status of each participant at the beginning of the trial to facilitate accurate predictions of trial outcomes."
  },
  "patient_performance_status": {
    "feature_name": "patient_performance_status",
    "feature_idea": "Categorical feature based on ECOG performance status, indicating health and functional status.",
    "feature_type": {
      "value": "categorical"
    },
    "data_sources": [
      "pubmed",
      "current_trial_summary",
      "related_clinical_trials"
    ],
\end{lstlisting}
\end{tcolorbox}
\end{figure*}

\begin{figure*}[hbt!]
\small
\begin{tcolorbox}[title=Example LLM Response (Phase I Plans) (cont'd), colback=gray!5, colframe=black, fonttitle=\bfseries]
\begin{lstlisting}[style=json]
    "example_values": [
      {
        "value": "0"
      },
      {
        "value": "1"
      },
      {
        "value": "2"
      },
      {
        "value": "3"
      },
      {
        "value": "4"
      },
      {
        "value": "5"
      }
    ],
    "possible_values": {
      "value": [
        "0",
        "1",
        "2",
        "3",
        "4",
        "5"
      ]
    },
    "feature_instructions": "Research and compile the ECOG performance status categories, which range from 0 to 5. Ensure that the feature captures the patient's performance status accurately at the beginning of the trial. The values should be represented as strings corresponding to the ECOG scale. This feature should be derived from patient records and trial documentation, ensuring consistency in how the performance status is recorded across different trials."
  },
  "trial_design_elements": {
    "feature_name": "trial_design_elements",
    "feature_idea": "Categorical feature indicating design elements like randomization and blinding that ensure reliability of trial outcomes.",
    "feature_type": {
      "trial_design_elements": "multi-categorical"
    },
    "data_sources": [
      "current_trial_summary",
      "pubmed"
    ],
    "example_values": [
      {
        "trial_design_elements": [
          "randomized",
          "double-blind"
        ]
      },
      {
        "trial_design_elements": [
          "non-randomized",
          "open-label"
        ]
      }
    ],
    "possible_values": {
      "trial_design_elements": [
        "randomized",
        "non-randomized",
        "double-blind",
        "single-blind",
        "open-label",
        "placebo-controlled"
      ]
    },
    "feature_instructions": "Research and compile a list of design elements commonly used in clinical trials that enhance the reliability of outcomes. Focus on elements such as randomization, blinding, and control groups. Ensure that the feature captures whether the trial is randomized or non-randomized, and whether it is blinded (single or double) or open-label. The final output should be a JSON object with a key \"trial_design_elements\" containing an array of applicable design elements for each trial."
  }
}

\end{lstlisting}
\end{tcolorbox}
\caption{Example prompt and response for the Feature Planner agent.}
\label{fig:feature_planner}
\end{figure*}

\begin{figure*}[hbt!]
\small
\begin{tcolorbox}[title=Feature Builder Prompt Instructions, colback=gray!5, colframe=black, fonttitle=\bfseries]

\textbf{Instructions (Researcher):} You are part of a clinical research team creating features for clinical trial machine learning models. You are investigating a particular clinical trial. You are given a dict of features that your team needs to do research on. You should make use of the given tools to do deep research, gather information and provide the data necessary to build all the features. Do not focus on formatting the features correctly, instead focus on making sure you have a full and complete set of data.

\textbf{Instructions (Builder):} You are part of a clinical research team creating features for clinical trial machine learning models. You are investigating a particular clinical trial. You are given a dict of features and their corresponding plans that your team needs to construct. A previous step has already gathered the necessary research results for these features, your job is to CORRECTLY construct these in the format prescribed by the feature plan. If there is
\begin{itemize}
    \item insufficient information
    \item missing information
    \item uncertainty/ambiguity
\end{itemize}
for any of the features, you should return the value 'None' for that feature (or sub-feature) and provide explanations for the feature you can't build. YOU MUST HAVE AN OUTPUT FOR EACH FEATURE.

\textbf{Input:} \begin{verbatim}
NCT ID: NCT01224639
Feature Plans: route_of_administration, dosing_regimen, previous_trial_success_rate
\end{verbatim}

\end{tcolorbox}

\vspace{1em}

\begin{tcolorbox}[title=Example LLM Response (Researcher), colback=gray!5, colframe=black, fonttitle=\bfseries]
\begin{lstlisting}[style=json]
1. **Route of Administration**: The trial NCT01224639 uses two routes of administration:
   - Subcutaneous (SC)
   - Intradermal (ID)

2. **Dosing Regimen**: The dosing regimens for the trial are categorized as follows:
   - Low Dose: Administered SC (0.5 mL) and ID (0.1 mL) on Days 0 and 90.
   - High Dose: Administered SC (0.5 mL) and ID (0.1 mL) on Days 0 and 90.
   - Placebo: Phosphate buffered saline administered SC (0.5 mL) and ID (0.1 mL).

3. **Previous Trial Success Rate**: 
   - Related trials in the dengue fever therapeutic area include:
     - NCT01134263: Phase 3 trial, outcome: Success
     - NCT00468858: Phase 2 trial, outcome: Success
     - NCT01110551: Phase 1 trial, outcome: Success
     - NCT00370682: Phase 2 trial, outcome: Success
     - NCT00617344: Phase 2 trial, outcome: Success
   - Total successful trials: 5
   - Total trials analyzed: 5
   - Previous trial success rate = 5 successful trials / 5 total trials = 1.0 (or 100%).

This information provides a comprehensive basis for building the features related to the route of administration, dosing regimen, and previous trial success rate.

\end{lstlisting}
\end{tcolorbox}

\begin{tcolorbox}[title=Example LLM Response (Builder), colback=gray!5, colframe=black, fonttitle=\bfseries]
\begin{lstlisting}[style=json]
{
  "feature_values": {
    "route_of_administration": {
      "route_of_administration": "subcutaneous"
    },
    "dosing_regimen": {
      "dosing_regimen": "multiple doses"
    },
    "previous_trial_success_rate": {
      "value": 1.0
    }
  }
}
\end{lstlisting}
\end{tcolorbox}

\caption{Example prompt and response for the Feature Builder agent.}
\label{fig:feature_builder}
\end{figure*}

\begin{figure*}[hbt!]
\small
\begin{tcolorbox}[title=Model-Based Evaluator Prompt Instructions, colback=gray!5, colframe=black, fonttitle=\bfseries]

\textbf{Role:} You are an experienced biomedical data scientist.

\textbf{Task:} You are supervising the construction of a machine learning model for a specific clinical trial task.

\textbf{Instructions:} The model must be built with features from data from ClinicalTrials.gov and from past scholarly research from PubMed.

A version of the model has been trained, and you are provided the current performance. Please provide suggestions for
\begin{itemize}
    \item additional features
    \item refinements to the existing features
    \item features to remove
\end{itemize}
Keep your suggestions concise, and limit to a maximum of 2-3 suggestions.

\textbf{Input:} \begin{verbatim}
roc_auc_score: 0.73
current_features_with_plan: … [feature plans]
feature_importances: … [feature names with coefficients]
\end{verbatim}
\end{tcolorbox}

\vspace{1em}

\begin{tcolorbox}[title=Example LLM Response (Model-Based Evaluator), colback=gray!5, colframe=black, fonttitle=\bfseries]
\begin{lstlisting}[style=json]
[
 "Consider adding a feature for 'historical trial outcomes' that captures the success rates of previous trials in the same therapeutic area, as this could provide valuable context for predicting current trial outcomes.",
 "Refine the 'intervention_type' feature by expanding the categories to include more specific types of interventions, as well as ensuring that the feature captures the nuances of combination therapies if applicable.",
 "Remove or replace the 'gender_inclusion' feature, as it currently does not contribute to the model's predictive power and may not provide significant insights into trial outcomes."
]
\end{lstlisting}
\end{tcolorbox}

\caption{Example prompt and response for the Model-Based Evaluator agent.}
\label{fig:model_based_evaluator}
\end{figure*}

\begin{figure*}[hbt!]
\small
\begin{tcolorbox}[title=Example-Based Evaluator Prompt Instructions, colback=gray!5, colframe=black, fonttitle=\bfseries]

\textbf{Role:} You are an experienced clinical researcher.

\textbf{Task:} You are supervising the construction of a machine learning model for a specific clinical trial task.

\textbf{Instructions:} The model must be built with features from data from ClinicalTrials.gov and from past scholarly research from PubMed.

A version of the model has been trained, and you are provided the current performance, and an example of an incorrect prediction from the current model.

Based on the example and using the tools provided to help with further research, please conduct some analysis on why the model made the incorrect prediction. You should consider
\begin{itemize}
    \item features that were missed, and could have helped with the prediction
    \item features that were not useful
    \item misconstructed features
    \item feature plans that are not properly set up (e.g. missing instructions / missing categories)
\end{itemize}
Your analysis should be generalizable to other trials where possible. Keep your analysis concise.

\textbf{Input:} \begin{verbatim}
roc_auc_score: 0.73
current_features_with_plan: … [feature plans]
feature_importances: … [feature names with coefficients]
example: 
        ## NCT02726334 Predicted 1, should be 0

        ### Features
        …. [features]

        ### Reasons for features that are None
        …
      "primary_outcome_measure": "The provided outcome measure 'maximum tolerated dose' does not fit into the predefined categories of safety, efficacy, pharmacokinetics, tolerability, or biomarkers."

\end{verbatim}
\end{tcolorbox}

\vspace{1em}

\begin{tcolorbox}[title=Example LLM Response (Example-Based Evaluator), colback=gray!5, colframe=black, fonttitle=\bfseries]
\begin{lstlisting}[style=json]
The analysis of the trial NCT02726334 reveals several key areas for improvement in the feature set used by the model. Firstly, the `primary_outcome_measure` should include specific measures relevant to the trial, such as "maximum tolerated dose," which was not categorized correctly. This misclassification can lead to significant inaccuracies in predicting trial outcomes. Secondly, the `trial_design_elements` feature must accurately reflect the trial\'s design, which in this case is non-randomized and single-group, rather than allowing for invalid values. The `number_of_participants` is a crucial feature that was missing, as it provides insight into the trial\'s power and potential for generalizability. Furthermore, the trial location should be expanded to include Australia, as limiting geographical categories can exclude relevant data. Overall, these findings suggest that careful attention to feature categorization and completeness is essential for improving the predictive accuracy of models in clinical trial outcomes.
\end{lstlisting}
\end{tcolorbox}

\caption{Example prompt and response for the Example-Based Evaluator agent.}
\label{fig:example_based_evaluator}
\end{figure*}

\end{document}